\documentclass[transmag]{IEEEtran}
\usepackage{latexsym}
\usepackage{graphicx}
\usepackage{amsfonts,amssymb,amsmath}
\usepackage{hyperref}
\usepackage{multirow}
\usepackage{float}
\usepackage{epsfig}
\usepackage{booktabs}
\usepackage{color}
\usepackage{threeparttable}
\usepackage{tabularx}
\usepackage{subfigure}
\usepackage{hyperref}
\hypersetup{hypertex=true,
            colorlinks=true,
            linkcolor=blue,
            anchorcolor=blue,
            citecolor=blue}
\usepackage[
    natbib=true,
    style=numeric,
    sorting=nty
]{biblatex}

\def\BibTeX{{\rm B\kern-.05em{\sc i\kern-.025em b}\kern-.08em T\kern-.1667em\lower.7ex\hbox{E}\kern-.125emX}}

\begin{document}

\title{PDWN: Pyramid Deformable Warping Network for Video Interpolation}

\author{Zhiqi Chen, Ran Wang, Haojie Liu and Yao Wang
%\IEEEmembership{Member, IEEE}
%\thanks{This paragraph of the first footnote will contain the 
%date on which you submitted your paper for review. It will also contain 
%support information, including sponsor and financial support acknowledgment. 
%For example, ``This work was supported in part by the U.S. Department of 
%Commerce under Grant BS123456''.}
%\thanks{The next few paragraphs should contain 
%the authors' current affiliations, including current address and e-mail. For 
%example, F. A. Author is with the National Institute of Standards and 
%Technology, Boulder, CO 80305 USA (e-mail: author@ boulder.nist.gov).}
%\thanks{S. B. Author, Jr., was with Rice University, Houston, TX 77005 USA. He is 
%now with the Department of Physics, Colorado State University, Fort Collins, 
%CO 80523 USA (e-mail: author@lamar.colostate.edu).}
%\thanks{T. C. Author is with 
%the Electrical Engineering Department, University of Colorado, Boulder, CO 
%80309 USA, on leave from the National Research Institute for Metals, 
%Tsukuba, Japan (e-mail: author@nrim.go.jp).}
}

\IEEEtitleabstractindextext{\begin{abstract} Video interpolation aims to generate a non-existent intermediate frame given the past and future frames. Many state-of-the-art methods achieve promising results by estimating the optical flow between the known frames and then generating the backward flows between the middle frame and the known frames. However, these methods usually suffer from the inaccuracy of estimated optical flows and require additional models or information to compensate for flow estimation errors. Following the recent development in using deformable convolution (DConv) for video interpolation, we propose a light but effective model, called Pyramid Deformable Warping Network (PDWN). PDWN uses a pyramid structure to generate DConv offsets of the unknown middle frame with respect to the known frames through coarse-to-fine successive refinements. Cost volumes between warped features are calculated at every pyramid level to help the offset inference. At the finest scale, the two warped frames are adaptively blended to generate the middle frame. Lastly, a context enhancement network further enhances the contextual detail of the final output. Ablation studies demonstrate the effectiveness of the coarse-to-fine offset refinement, cost volumes, and DConv. Our method achieves better or on-par accuracy compared to state-of-the-art models on multiple datasets while the number of model parameters and the inference time are substantially less than previous models. Moreover, we present an extension of the proposed framework to use four input frames, which can achieve significant improvement over using only two input frames, with only a slight increase in the model size and inference time.
\end{abstract}

%(Note that the organization of the body of the paper is at the authors' 
%discretion; the only required sections are Introduction, Methods and 
%Procedures, Results, Conclusion, and References. Acknowledgements and 
%Appendices are encouraged but optional.)

\begin{IEEEkeywords}
Video interpolation, deformable convolution, deep learning
%At least four keywords or phrases in 
%alphabetical order, separated by commas. For a list of suggested keywords, 
%send a blank e-mail to \href{mailto:keywords@ieee.org}{mailto:keywords@ieee.org} or visit 
%\href{http://www.ieee.org/organizations/pubs/ani_prod/keywrd98.txt}{http://www.ieee.org/organizations/pubs/ani\_prod/keywrd98.txt}
\end{IEEEkeywords}

%Note: There should no nonstandard abbreviations, acknowledgments of support, 
%references or footnotes in in the abstract.
}

\maketitle

\section{INTRODUCTION}
Video interpolation, which aims to generate intermediate frames between given prior (or left) and post (or right) frames, is widely applied in video coding \cite{wu2018video} and video frame rate conversion \cite{castagno1996method}. However, natural videos include complicated appearance and motion dynamics, e.g., various object scales, different viewpoints, varied motion patterns, object occlusions, and dis-occlusions, making interpolation of realistic frames a significant challenge.
\begin{figure*}[tb]
\centering
    \subfigure[Accuracy / size tradeoff]{
        \includegraphics[width=.48\textwidth]{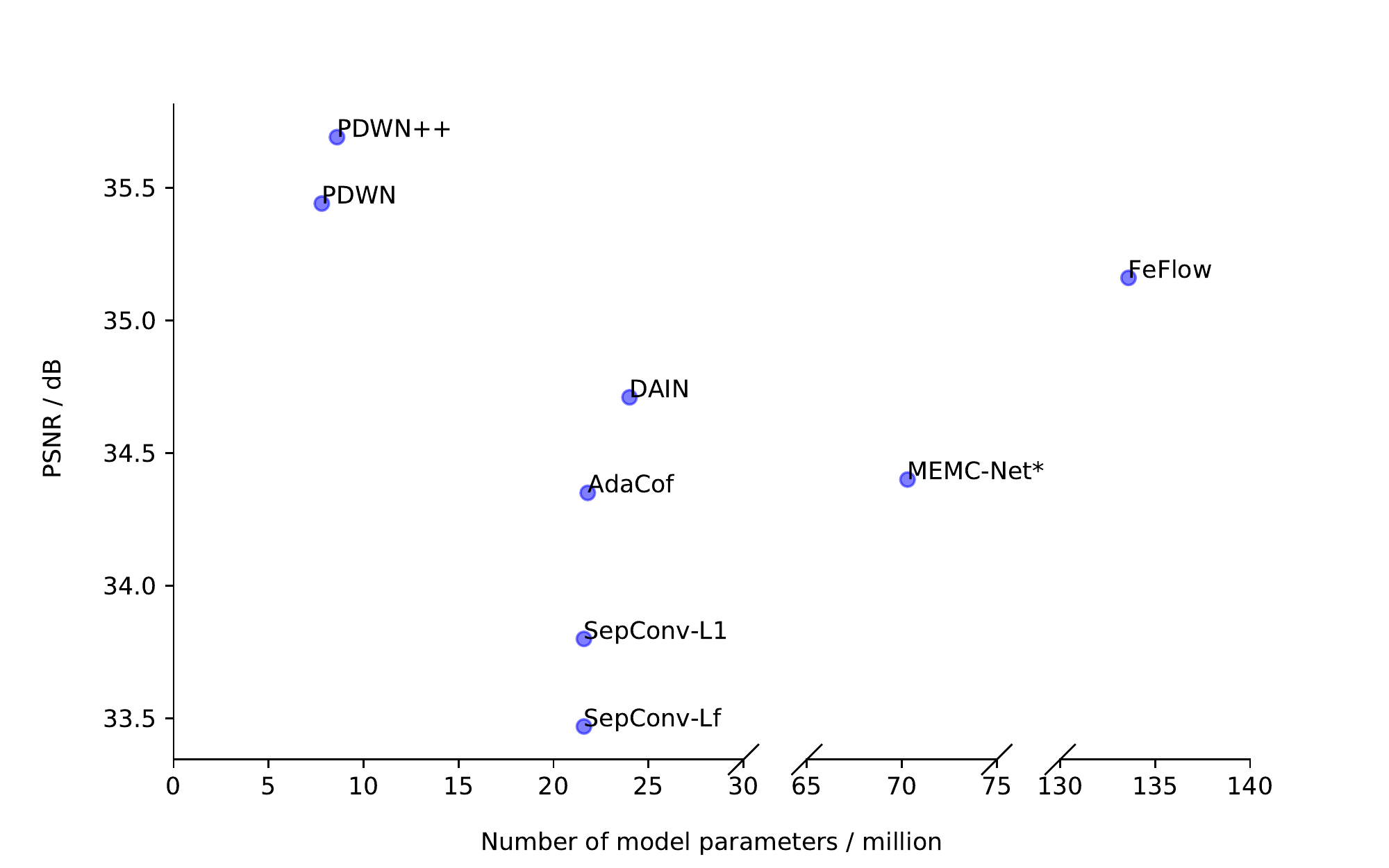}}
    \subfigure[Accuracy / runtime tradeoff]{
        \includegraphics[width=.48\textwidth]{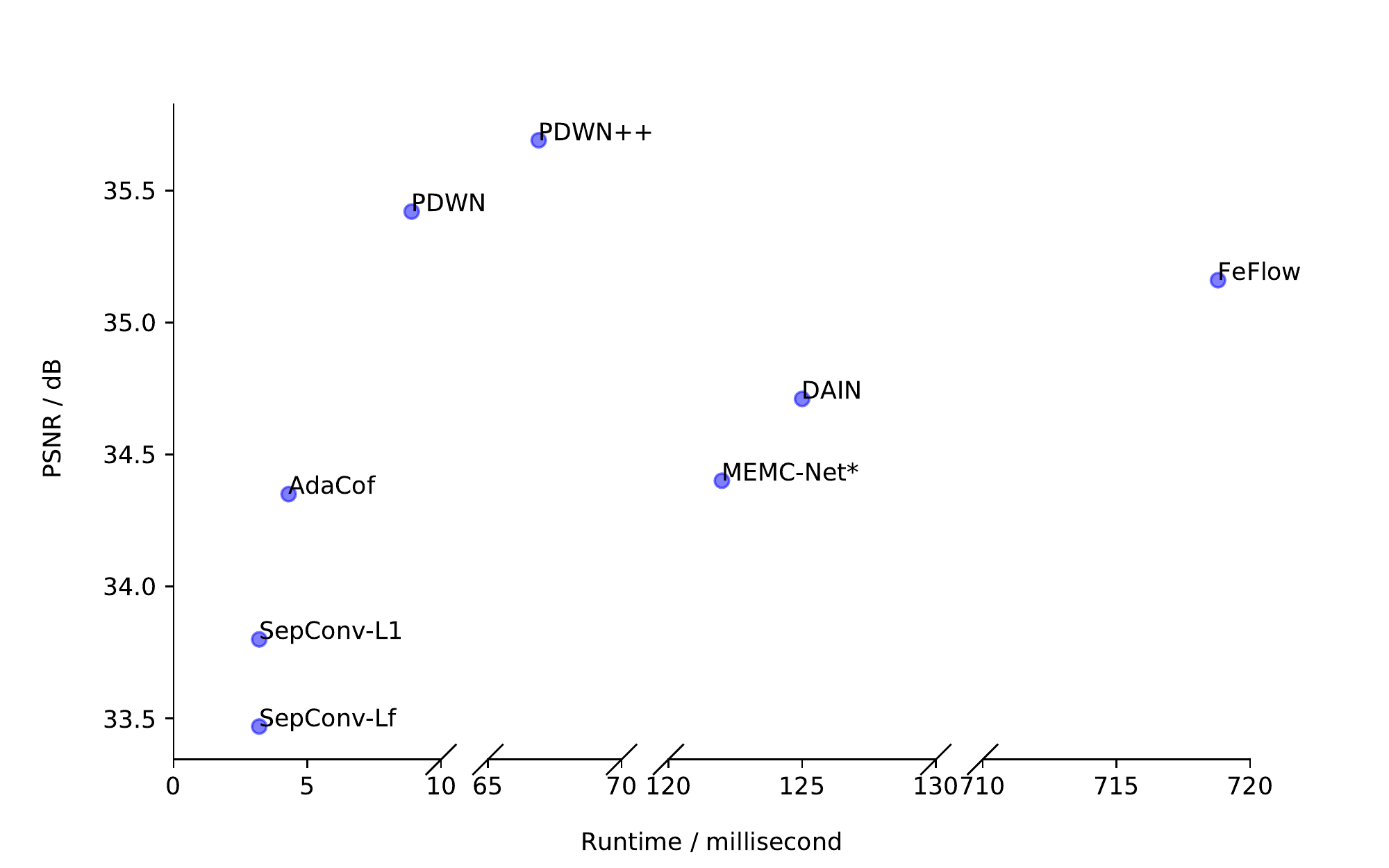}}
\caption{{\bf Accuracy / efficiency tradeoff for video interpolation on Vimeo-90K dataset:} PDWN in comparison to previous work. Left: PDWN outperforms state-of-the-art methods in both accuracy and model size. Right: PDWN reaches the best balance between accuracy and runtime. PDWN++ is the enhanced PDWN model with input normalization, network improvements, and self-ensembling. PDWN++ further improves the performance with a small cost of model size and nearly 8 times of the runtime. The runtime is the time needed to interpolate one frame on GeForce RTX 2080 Ti GPU card.}
\label{fig:size}
\end{figure*}

Flow-based methods have been proven to work well in video interpolation \cite{bao2019depth,bao2019memc,jiang2018super,liu2017video,xu2019quadratic}. Many state-of-the-art methods first use an optical flow estimator to obtain optical flow between given frames, and then infer the optical flow between the missing middle frame and the left and right known frames, respectively, by prefixed motion assumptions such as linear motion \cite{bao2019depth,jiang2018super,Niklaus_CVPR_2020} or quadratic motion \cite{xu2019quadratic}. The middle frame is then obtained by backward warping input frames using the estimated optical flows. Such approaches are prone to flow errors caused by adopted flow estimators and the errors in the motion assumption. Thus, additional flow correction networks\cite{xu2019quadratic} or additional information such as depth \cite{bao2019depth} are usually required to refine the initial interpolated optical flows, leading to sophisticated models. Moreover, training such models require ground truth optical flow or depth information, which is expensive to obtain in large quantities.

Though flow-based methods have achieved great success in video interpolation, they are prone to errors and face the challenge of complicated dynamic scenes including nonlinear motions, lighting changes, and occlusions. Recently, deformable convolution (DConv) have been investigated in video interpolation to warp features and frames \cite{gui2020featureflow,lee2020adacof}. DConv produces multiple offsets for each pixel to be interpolated with respect to each input frame, and uses a weighted average of these offset pixels in the previous (or future) frame to predict target pixel. When the filter size of DConv is 1x1 and the filter coefficient is 1, DConv offset is the same as optical flow. When the filter size is larger than 1, DConv performs many-to-one weighted warping and thus the offsets can be considered as many-to-one flows. Generally, Dconv offsets are more robust than single optical flow. Furthermore, DConv filter coefficients enable the model to produce more complex transformations. However, the increased degree of freedom of DConv makes the model hard to train.

To alleviate the above issues, we propose a pyramid deformable warping network (PDWN) to perform coarse-to-fine frame warping. The coarse-to-fine structure has been proved to be powerful in optical flow estimation \cite{ilg2017flownet,ranjan2017optical,sun2018pwc}. In video interpolation, however, relatively few approaches explored the coarse-to-fine strategy. Amersfoort and Shi \cite{van2017frame} proposed a multi-scale generative adversarial network to generate the predicted flow and the synthesized frame in a coarse-to-fine fashion. Zhang et. al. \cite{zhang2020flexible} designed a recurrent residual pyramid architecture to refine optical flow using a shared network across pyramid levels. Other methods, despite the usage of multi-scale features, only generate one-stage optical flow \cite{bao2019depth,gui2020featureflow,liu2017video}. In our work, we exploit the advantages of the warping strategy and cost volume in addition to the pyramid structure to estimate DConv offsets from coarse to fine. 

The proposed network follows a pyramid structure that extracts features at various resolution scales from each input frame. At every pyramid level, DConv is adopted to warp features from the past and future frames towards the middle frame, and a matching cost volume under different additional offsets between two warped features is constructed and exploited to infer residual DConv offsets. By warping features with the obtained offsets and passing the cost volume to the next pyramid level, the network refines the estimated offsets from coarse to fine. We demonstrate that such a methodology for video interpolation generates more realistic frames without requiring additional information such as ground truth optical flow information or depth during training. Our proposed network greatly reduces the number of model parameters and the inference time, while achieving better or on-par performance compared to state-of-the-art models as shown in Figure \ref{fig:size}. Furthermore, our proposed approach can be extended to using multiple input frames easily, and using four instead of two frames as input leads to significantly improved interpolation results.

\section{Related Work}
\subsection{Video interpolation} Video interpolation has been extensively explored in the literature \cite{bao2019depth,bao2019memc,gui2020featureflow,jiang2018super,lee2020adacof,liu2017video,Niklaus_CVPR_2020,niklaus2017video,xu2019quadratic,xue2019video}. Prior methods can be grouped into two categories: kernel-based approach and flow-based approach. Kernel-based approaches \cite{lee2020adacof,niklaus2017avideo,niklaus2017video} estimate convolution kernel parameters to hallucinate intermediate frame. However, kernel-based approaches typically fail in cases with large motions unless very large filter kernels are used, and suffer from large computational loads. Flow-based approaches estimate the optical flow to warp pixels to synthesize the target frame. Super SloMo \cite{jiang2018super} adopted one UNet to estimate optical flow between two input frames, and another UNet to correct the linearly interpolated flow vector. Beyond linear motion assumptions, Quadratic flow (QuaFlow) \cite{xu2019quadratic} adopted PWC-Net\cite{sun2018pwc} to estimate optical flow between input frames. Then the quadratically interpolated flow was refined through a UNet. MEMC-Net\cite{bao2019memc} estimated both motion vectors and compensation filters through CNNs. Note that four input frames are required for QuaFlow to construct a quadratic model. Instead of bilinear interpolation, MEMC proposed an adaptive warping layer based on optical flow and compensation filters to reduce blur. Based on MEMC-Net, DAIN \cite{bao2019depth} used depth information estimated by a pre-trained hourglass architecture \cite{li2018megadepth} to detect occlusions. Different from above methods, Softmax Splatting \cite{Niklaus_CVPR_2020} estimated forward flow using an off-the-shelf optical flow estimator and designed a differentiable way to do forward warping. Though flow-based methods can generate sharp frames, inaccurate flow estimation often leads to severe artifacts. Unlike methods described above, our method directly estimates the "flow" between given input frames and the unknown middle frame without assuming the trajectory is linear or quadratic or has other parametric forms. And we estimate many-to-one "flow" which is more robust compared to single optical flow. Furthermore, we estimate the flows in a coarse-to-fine manner, to efficiently handle large motions.

\subsection{Pyramid structure and the cost volume} Pyramid structure has been proved to be powerful in optical flow estimation. Ilg et al. \cite{ilg2017flownet} achieved state-of-the-art performance by stacking several UNets into a large model, called FlowNet2. To reduce over-fitting problem caused by large models, SpyNet \cite{ranjan2017optical} incorporated two classical principles, pyramid structure, and warping, into deep learning. A spatial pyramid network was constructed for each of the two frames, and it estimated the flow in each scale and warped the second image to the first one at each scale repeatedly to reduce motion between two images. PWC-Net \cite{sun2018pwc} further explored the trade-off between accuracy and model size. Instead of image pyramids, PWC-Net constructed feature pyramids that are invariant to shadows and lighting change. Partial cost volume is used to represent matching cost associated with different disparities. Inspired by classical pyramid energy minimization in optical flow algorithms, RRPN \cite{zhang2020flexible} designed a recurrent residual pyramid architecture for video frame interpolation to refine optical flow using a shared network for every pyramid level. Following above methods, we also exploit the advantages of classical principles of optical flow -- the pyramid structure, multi-scale warping, and cost volume. Different from RRPN, we replace the flow estimation in each scale by the estimation of many-to-one offset maps through the use of deformable convolution filters, significantly reducing artifacts that are associated with occasional wrong flow estimates. Furthermore, cost volume is incorporated into our model non-trivially. We demonstrate that cost volume between the warped features of the two known frames can provide useful information for estimating the flow between the unknown middle frame and the known prior and post frames. 
%Instead of estimating a single optical flow between two known frames, we estimate two sets of many-to-one flows, between the unknown middle frame and the known prior and post frames in a coarse to fine manner.

\subsection{Deformable convolution} DConv operation \cite{dai2017deformable} is originally proposed to overcome the limitation of CNNs due to fixed filter support configuration and to enhance the transformation modeling capacity of CNNs. It estimates a set of $K$ offsets at each pixel and a global filter (non-spatially varying) with $K$ coefficients to be applied for the $K$ offset pixels. Zhu et. al. \cite{zhu2018deformable} further improved DConv by adding spatially adaptive modulation weights to modulate the global filter coefficient associated with each offset. The improved DConv thus has the ability to vary the attention to different offset pixels. Recognizing that DConv can be viewed as many-to-one weighted backward warping, FeFlow \cite{gui2020featureflow} used DConv to align input features from two known frames and fused aligned features to synthesize the middle frames. AdaCoF \cite{lee2020adacof} constructed a UNet to estimate both local filter weights and offsets for each target pixel to synthesize output frames. We have found that learning a global filter plus spatially-varying modulation weights as in \cite{zhu2018deformable} is better than directly estimating locally adaptive filters. Different from FeFlow and AdaCoF \cite{lee2020adacof} that estimate the DConv offsets directly in the original image resolution, we perform offset estimation and feature alignment in a coarse-to-fine successive refinement manner. Specifically, we successively refine DConv offsets from the coarser scales to the finer scales. We further utilize the cost volume computed from two aligned features at each scale to improve the accuracy of the offset update. 

\section{Methods}
\begin{figure*}[htp]
\begin{center}
\includegraphics[width=\linewidth]{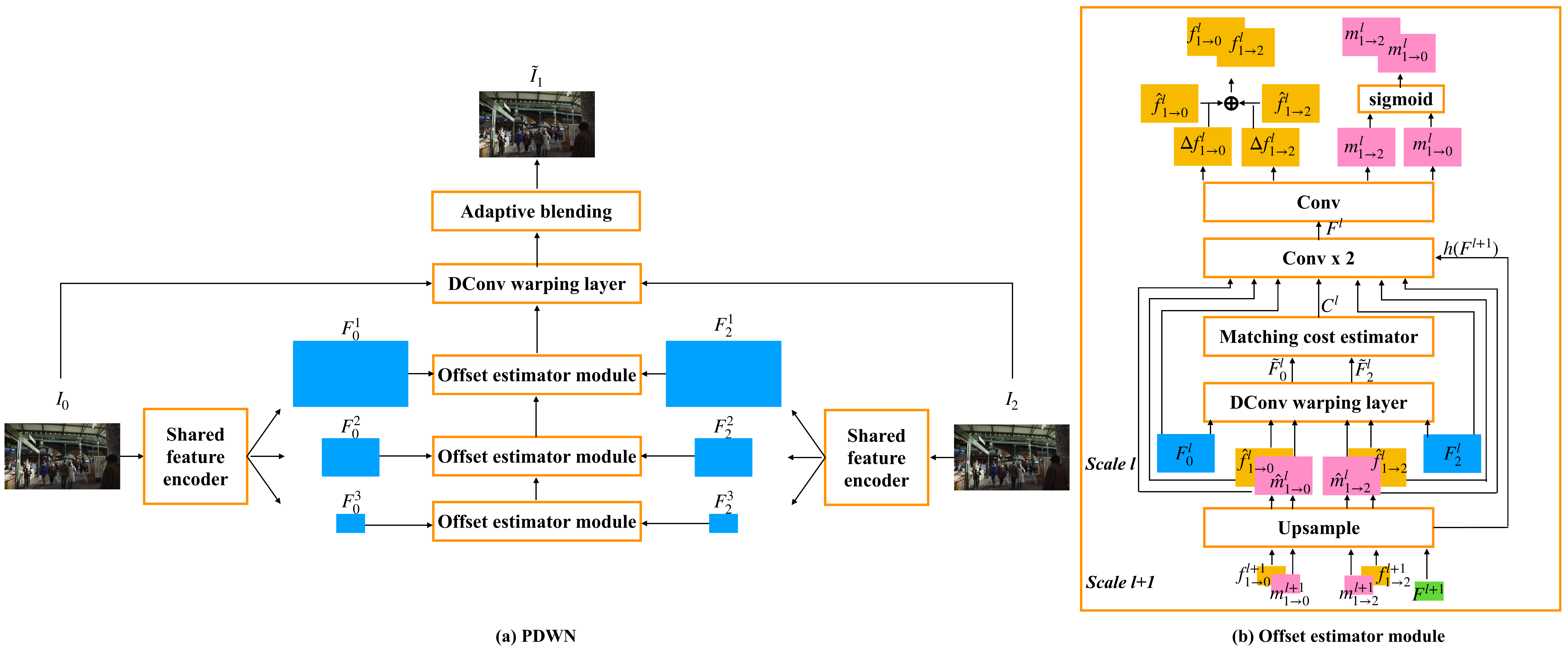}
\end{center}
   \caption{{\bf (a) The architecture of PDWN}. Given the past frame $I_0$ and the future frame $I_2$, PDWN first generates two feature pyramids. Then DConv offsets $f_{1\to0}^l, f_{1\to2}^l$ and associated modulation weights $m_{1\to0}^l, m_{1\to2}^l$ are generated from the coarsest scale to the finest scale. Finally, two warped frames are adaptively blended to synthesize the middle frame. {\bf (b) Offset estimator module}. Features of scale $l$ are warped producing $\tilde{F_0^l}$ and $\tilde{F_2^l}$ via DConv with generated offsets and associated modulation weights. The cost volume between $\tilde{F_0^l}$ and $\tilde{F_2^l}$, together with input features, are fed to 2 convolutional layers to refine the next-scale DConv offsets and the associated modulation weights. The above process is repeated until it gets to the finest level.}
\label{fig:network_structure}
\end{figure*}
The structure of PDWN is shown in Figure \ref{fig:network_structure}. Given two input frames $I_0$ and $I_2$, we aim to synthesize the intermediate frame $I_1$ by gradually warping features of input frames to the middle frame using DConv. First, we construct a feature pyramid for each input frame using a shared feature extractor. Second, we generate the offsets and the associated modulation weights between each input frame and the middle frame, and then warp features of both input images toward the middle frame. This operation is taken at every level of the feature pyramid to refine the motion. Thus, the estimated DConv offsets, which can be considered as many-to-one flow, are refined from the coarse level to the fine level. Third, at the finest resolution level (same as the input frame), interpolation weight maps between the warped left and right frames are generated to handle occlusions. Finally, following the post-processing scheme of DAIN, we adopt a context enhance network to further enhance the interpolated frame, shown in Figure \ref{fig:res}.

\subsection{Shared feature pyramid encoder} 
A multi-layer CNN is used to construct $L$-scale pyramids of feature representations for both input frames $\{F_i^l \mid i\in\{0,2\}, l\in\{1,2,...,L\}$. The features at the first scale, $F_i^1$,  have the same spatial resolution as the input frames. The $l$th scale feature $F_i^l$ is downsampled by a factor of 2 both horizontally and vertically from the $(l-1)$-th scale feature $F_i^{l-1}$. Each scale consists of two convolution blocks, with specifics shown in Table \ref{table:configuration}.

\subsection{Offset estimator module} 
The Offset estimator module is used in every scale of PDWN. It jointly predicts the DConv offsets from the unknown intermediate frame to the given input frames and the associated modulation weights for each offset in order to warp input frames and features to the intermediate frame. 

\paragraph{Deformable warping with spatially-varying modulation coefficients} A deformable convolution filter is specified by a global filter $w(j)$, a set of spatially-varying offsets $f(j, x)$, and modulation coefficients $m(j, x)$, where $j$ denotes $j$-th location in a filter support $\mathcal{R}$ and $x$ indicates pixel location. The global filter $w(j)$ here is the same convolution filter as regular convolutions except that the sampling is irregular. The support $\mathcal{R}= \{(-1, -1), (-1, 0), ..., (0, 1), (1, 1)\}$ specifies a $3\times 3$ filter in our model. The offset is defined by horizontal and vertical displacements. And every sampling point is associated with a modulation weight. Thus the offset tensor and the modulation tensor have channel dimensions of 18 and 9, respectively. The global filter has the size of $3\times 3$. To use DConv for video interpolation at multiple scales, we generate two sets of offsets and modulation coefficients at scale $l$, ${f_{1\to i} ^l(j, x)}, {m_{1\to i}^l (j, x)} $ with $i=0,2$ indicating the known prior and post frame and $i=1$ the unknown middle frame. The global filter weights $w^l(j)$ are learnt and stay fixed after training for each scale and shared for known input features. Specifically, we generate the warped feature at scale $l$ at pixel $x$ from the original features for frame i as follows:
\begin{equation}
\tilde{F}_i^l(x) = \sum_{j=1}^{\left| R\right|} w^l(j)m_{1\to i}^l(j, x)F_{i}^{l}(x+R(j)+ f_{1\to i}^l(j, x))
\label{align}
\end{equation}

\paragraph{Cost volume between features warped towards the middle frame} The notion of cost volumes has been widely used in optical flow methods \cite{hosni2012fast,sun2018pwc,vzbontar2016stereo} to provide explicit representation of matching cost under different displacements between two given frames for each pixel. In the PWC method for optical flow estimation, the cost volume is constructed between a warped image and a fixed image. Typically, for each pixel  $x$ in one frame, the correlation between the feature at $x$ in this frame and the feature at a displaced location $x+d$ in the other frame is computed, for a finite set of displacements $d\in {\mathcal{D}_k(x)}$. $\mathcal{D}_k(x)$ is a square neighborhood of pixel $x$ with neighborhood size $k\times k$. In our case, however, a cost volume is calculated between two sets of warped features $\tilde F_0^l$ and $\tilde F_2^l$ based on the estimated offsets from each known frame to the middle frame, determined in a lower scale. The cost volume indicates the correlation between the features for corresponding pixels in the left and the right warped features under different displacements. Specifically, given $\tilde{F}_{0}^{l}$ and $\tilde{F}_{2}^{l}$, a cost volume $C^l$ is constructed based on
\begin{equation}
C^l(x_1, x_2) = \frac{1}{k^2}\tilde{F}_0^l(x_1)^T\tilde{F}_2^l(x_2), \quad x_2 \in \mathcal{D}_k(x_1)
\label{cost}
\end{equation}
where $x_1$ and $x_2$ are pixel indexes. We set $k=9$, including displacement from -4 to 4 in both horizontal and vertical directions. Thus the cost volume has a channel dimension of 81.

Instead of using a pre-determined way to calculate the matching cost, one can also train a small network (learnt as part of the entire network) $v(\cdot)$ that takes the two warped features and outputs the cost volume:
\begin{equation}
C^l = v(\tilde{F}_0^l, \tilde{F}_2^l)
\end{equation}
We experimented with both approaches, where we used a network with two conv layers for the network $v(\cdot)$.

\paragraph{Multi-scale offset estimation} As shown in Figure \ref{fig:network_structure}, we estimate the offsets between the middle frame and each of the two input frames from coarse to fine scales with a total of $L$ scales ($L=3$ in Figure \ref{fig:network_structure}). DConv offsets are generated within each scale to gradually reduce the distance between two sets of features warped towards the middle frame. 

At $l$-th scale, the offset estimation block first upsamples the estimated offsets $f^{l+1}_{1\to i}$ and modulation weights $m^{l+1}_{1\to i}$ at the lower scale $l+1$ to the current resolution using bilinear interpolater $h(\cdot)$, yielding
\begin{equation}
\hat{f}^l_{1\to i} = 2*h(f^{l+1}_{1\to i})
\end{equation}
\begin{equation}
    \hat{m}^l_{1\to i} = h(m^{l+1}_{1\to i})
\end{equation}
Then it warps the original features $F_i^l$ towards the middle frame based on $\hat{f}_{1\to i}^{l}$, $\hat{m}_{1\to i}^{l}$, and the learnt global filter $w^l$, generating the warped features $\tilde F_i^l$ using Eq. (\ref{align}). Then, the offset estimator computes the cost volume $C^l$ between the two warped features using Eq. (\ref{cost}). Next, it generates two sets of DConv offsets residuals $\Delta f_{1\to i}^{l}$ and two sets of modulation weight $m_{1\to i}^{l}$ from $C^l$, $\hat{f}_{1\to i}^{l}$, $\hat{m}_{1\to i}^{l}$, the original features $F_{i}^l$, and the upsampled features 
$h(F^{l+1})$ from the features $F^{l+1}$ generated by the offset estimator in the previous scale:
\begin{equation}
\begin{aligned}
   \Delta f_{1\to i}^{l}, m_{1\to i}^{l} = g(C^l, F_{i}^{l}, \hat{f}_{1\to i}^{l}, \hat{m}_{1\to i}^{l}, h(F^{l+1})), i=0,2
\end{aligned}
\label{warp1}
\end{equation}
where $g(\cdot)$ denotes a three-layer CNN. The final offsets and modulation weights are obtained by 
\begin{equation}
f_{1\to i}^{l} = \hat{f}_{1\to i}^{l} + \Delta f_{1\to i}^{l}
\label{warp2}
\end{equation}
\begin{equation}
% \{m_{1\to i}^{l}\}= sigmoid( h(\{m_{1\to i}^{l+1}\}) + \{\Delta m_{1\to i}^{l}\})
m_{1\to i}^{l}= \sigma(m_{1\to i}^{l})
\label{warp3}
\end{equation}
\begin{equation}
\sigma(t) = \frac{1}{1+e^{-t}}
\label{warp4}
\end{equation}
where $\sigma(\cdot)$ denotes a sigmoid activation function. We can use a small subnetwork (consisting of three conv layers) to estimate the offset fields because the motion between two warped features is usually small. The same process repeats until we complete scale 1. 

For the coarsest scale $L$, the offset estimator only takes the original features in that scale $F_0^{L}$ and $F_2^L$ as input and generates $f^L_{1\to i}$ and $m^L_{1\to i}$ directly.

To summarize, the offset estimator at each scale needs to generate two sets of offset tensors and two sets of modulation tensors, with a total channel dimension of 54. See Table \ref{table:configuration} for the specifics of the network structure. 

\subsection{Adaptive frame blending} 
Using the estimated offset $f_{1\to i}^1 $, modulation weights $m_{1\to i}^1$, and global filter $w^1$ at scale 1, we  warp frame $i$ towards the middle frame, generating two candidates estimates of the middle frames $\tilde{I}_i, i\in{0,2}$. Occlusions often happen due to the movement of objects. Therefore, in order to select valid pixels from two warped reference frames, we design a blending layer that generates a weight map $\alpha(x)$ to average the two transformed frames at position $x$. The layer is constructed by a three-layer CNN. See Table I, the network takes two warped frames, $\tilde{I}_0$ and $\tilde{I}_2$, and two warped features, $\tilde{F}_0^1$ and $\tilde{F}_2^1$, at first scale of the feature pyramids as input and generates the weight map with a softmax activation applied on the output layer. At position $x$, the blended frame is 
\begin{equation}
\tilde{I}_1(x) = \alpha(x) * \tilde{I}_0(x) + (1-\alpha(x)) * \tilde{I}_2(x)
\end{equation}
The warped features provide contextual information to estimate the weight map.

\subsection{Context enhancement network} 
\begin{figure}[htb]
\begin{center}
\includegraphics[width=\linewidth]{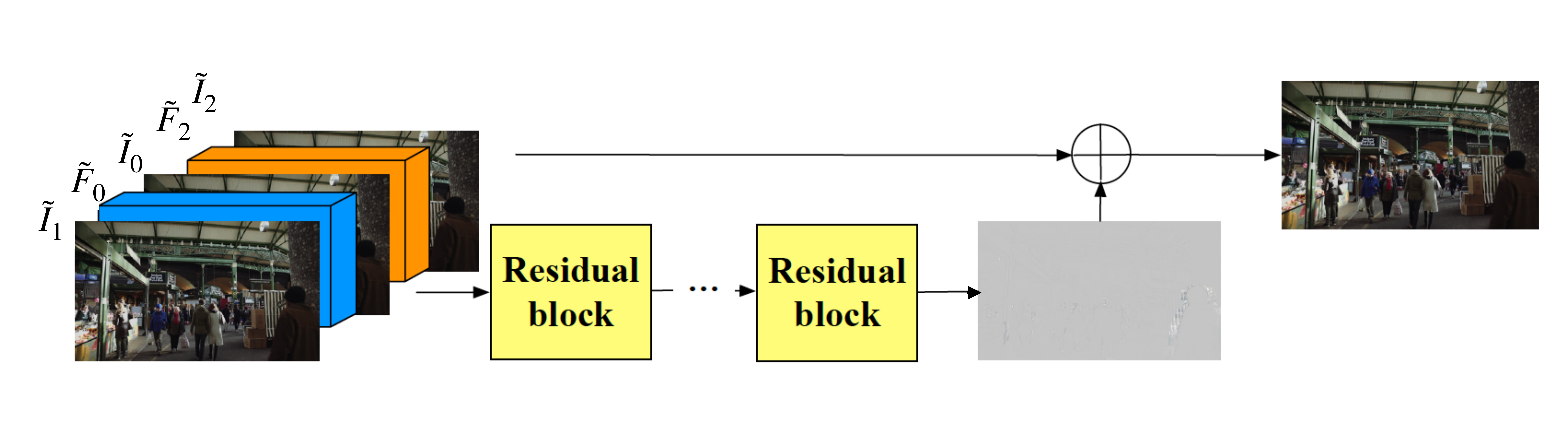}
\end{center}
   \caption{{\bf Context enhancement network.} After getting the initial synthesized middle frame, warped input frames and warped 1th-level features are fed into five residual blocks to further enhance contextual details of the synthesized frame.}
\label{fig:res}
\end{figure}

To generate the final output, we construct a context enhancement network which takes warped images and features at scale 1 as input and outputs a residual image between the unknown ground truth intermediate frame and the blended frame. The network consists of five residual blocks, shown in Figure \ref{fig:res}.  See Table \ref{table:configuration} for the specific network configuration.

\subsection{Extending to four input frames}
Quadratic flow \cite{xu2019quadratic} shows an improvement on moving trajectory estimation by estimating acceleration information from four input frames. We also extend our model to exploit the information in additional input frames and to estimate the motion more accurately. Our extended model takes four input frames (two previous and two following frames). A pyramid feature encoder is shared between four input frames to generate four feature pyramids. In the offset estimator, we input four feature maps of the four input frames instead of two in the first conv layer in Figure \ref{fig:network_structure}.(b). This allows the network to recognize the motion trajectory over a longer temporal scope and yield more accurate offset estimation. In higher scales, we still generate the warped feature maps for two closest past and future frames using the estimated offsets and modulation weights from the lower scale and determine the cost volume from these two warped features. Then the cost volume is concatenated with four original features of input frames as well as offsets and modulation weights and fed into the next scale to refine offsets and modulation weights in the next scale. Note that even though the input consists of four frames, the network only generates two sets of offsets, between the middle frame and its left and right neighboring frames, respectively. The final interpolated frame is the adaptively weighted average of these two closest frames warped by deformable convolution.

\subsection{Implementation detail}
\begin{table}
\caption{Architecture of PDWN}
\begin{center}
\renewcommand\tabcolsep{1pt}
\begin{threeparttable}
\begin{tabular}{cccc}
\toprule

Module & Scale & Output size & Configuration\\
\midrule
\multirow{13}{*}{Feature extractor} & \multirow{2}{*}{1} & \multirow{2}{*}{$H \times W$} & Conv 7 - 3 - 16 \\
\specialrule{0em}{1pt}{1pt}
  & & & Conv 5 - 16 - 16 \\
\cmidrule(r){2-4}
& \multirow{2}{*}{2} & \multirow{2}{*}{$H/2 \times W/2$} & Conv 3 - 16 - 32\\
\specialrule{0em}{1pt}{1pt}
 & & & Conv 3 - 32 - 32\\
\cmidrule(r){2-4}
& \multirow{2}{*}{3} & \multirow{2}{*}{$H/4 \times W/4$} & Conv 3 - 32 - 64\\
\specialrule{0em}{1pt}{1pt}
 & & & Conv 3 - 64 - 64\\
\cmidrule(r){2-4}
& \multirow{2}{*}{4} & \multirow{2}{*}{$H/8 \times W/8$} & Conv 3 - 64 - 96\\
\specialrule{0em}{1pt}{1pt}
 & & & Conv 3 - 96 -96\\
\cmidrule(r){2-4}
& \multirow{2}{*}{5} & \multirow{2}{*}{$H/16 \times W/16$}& Conv 3 - 96 - 128\\
\specialrule{0em}{1pt}{1pt}
 & & & Conv 3 - 128 - 128\\
\cmidrule(r){2-4}
& \multirow{2}{*}{6} & \multirow{2}{*}{$H/32 \times W/32$} & Conv 3 - 128 - 196\\
 \specialrule{0em}{1pt}{1pt}
 & & & Conv 3 - 196 - 196\\
\cmidrule(r){1-4}
  \multirow{24}{*}{Offset estimator} & \multirow{3}{*}{6} & \multirow{3}{*}{$H/32 \times W/32$} & Conv 3 - 473 - 256 \\
  \specialrule{0em}{1pt}{1pt}
 & & & Conv 3 - 256 - 256 \\
 \specialrule{0em}{1pt}{1pt}
 & & & Conv 3 - 256 - 54 \\
\cmidrule(r){2-4}
  & \multirow{4}{*}{5} & \multirow{4}{*}{$H/16 \times W/16$} & DConv 3 - 128 - 128 \\
  \specialrule{0em}{1pt}{1pt}
  & & & Conv 3 - 647 - 196 \\
  \specialrule{0em}{1pt}{1pt}
   & & & Conv 3 - 196 - 196 \\
   \specialrule{0em}{1pt}{1pt}
    & & & Conv 3 - 196 - 54 \\
\cmidrule(r){2-4}
  & \multirow{4}{*}{4} & \multirow{4}{*}{$H/8 \times W/8$} & DConv 3 - 96 - 96\\
  \specialrule{0em}{1pt}{1pt}
  & & & Conv 3 - 523 - 128 \\
  \specialrule{0em}{1pt}{1pt}
   & & & Conv 3 - 128 - 128 \\
   \specialrule{0em}{1pt}{1pt}
   & & & Conv 3 - 128 - 54 \\
\cmidrule(r){2-4}
  & \multirow{4}{*}{3} & \multirow{4}{*}{$H/4 \times W/4$} & DConv 3 - 64 - 64\\
  \specialrule{0em}{1pt}{1pt}
  & & & Conv 3 - 391 - 64 \\
  \specialrule{0em}{1pt}{1pt}
  & & & Conv 3 - 64 - 64\\
  \specialrule{0em}{1pt}{1pt}
   & & & Conv 3 - 64 - 54 \\
 \cmidrule(r){2-4}
  & \multirow{4}{*}{2} & \multirow{4}{*}{$H/2 \times W/2$} & DConv 3 - 32 - 32\\
  \specialrule{0em}{1pt}{1pt}
  & & & Conv 3 - 295 - 64 \\
  \specialrule{0em}{1pt}{1pt}
  & & & Conv 3- 64 - 64\\
  \specialrule{0em}{1pt}{1pt}
  & & & Conv 3 - 64 - 54 \\
 \cmidrule(r){2-4}
  & \multirow{4}{*}{1} & \multirow{4}{*}{$H \times W$} & DConv 3 - 16 - 16 \\
  \specialrule{0em}{1pt}{1pt}
  & & & Conv 3 - 231 - 64 \\
  \specialrule{0em}{1pt}{1pt}
  & & & Conv 3 - 64 - 64\\
  \specialrule{0em}{1pt}{1pt}
  & & & Conv 3 - 64 - 54 \\
\cmidrule(r){1-4}
 \multirow{5}{*}{Adaptive frame blending} &  \multirow{5}{*}{1} & \multirow{5}{*}{$H \times W$} & DConv 3 - 3 - 3 \\
 \specialrule{0em}{1pt}{1pt}
 & & & DConv 3 - 16 - 16\\
 \specialrule{0em}{1pt}{1pt}
 & & & Conv 3 - 38 - 16 \\ 
 \specialrule{0em}{1pt}{1pt}
 & & & Conv 3 - 16 - 16 \\
 \specialrule{0em}{1pt}{1pt}
 & & & Conv 3 - 16 - 2 \\
\cmidrule(r){1-4}
 \multirow{3}{*}{Context enhancement} & \multirow{3}{*}{1} & \multirow{3}{*}{$H \times W$} & Conv 3 - 41 - 64 \\
 \specialrule{0em}{1pt}{1pt}
 & & & Conv 3 - 64 - 64 $\times$ 2 $\times$ 4\\
\specialrule{0em}{1pt}{1pt}
  & & & Conv 3 - 64 - 3\\
\bottomrule
\end{tabular}

\begin{tablenotes}
        \footnotesize
        \item[*]  The convolutional and deformable convolutional layer parameters are denoted as “Conv/DConv $<$filter size$>$ - $<$number of input channels$>$ - $<$number of output channels$>$”. The leakyReLU activation function, max pool layer, bilinear upsample layer, and matching cost layer are not shown for brevity.

      \end{tablenotes}
\end{threeparttable}
\end{center} 
\label{table:configuration}
\end{table}
\paragraph{Architecture configurations}
The configurations of PDWN with 6 scales and predefined matching cost calculation, evaluated in this paper, are outlined in Table \ref{table:configuration}.

\paragraph{Loss function} L1 norm has been proven to generate less blurry results in image synthesis tasks \cite{DBLP:journals/corr/GoroshinML15,mathieu2015deep}. Thus, L1 Reconstruction loss between the reconstructed frame and the ground truth frame is used to train the model:
\begin{equation}
    \mathcal{L} = || \tilde{I}_1 - I_1 ||_1
\end{equation}
We also explore a multi-scale L1 reconstruction loss for training. Specifically, we downsample the input frames and the ground truth middle frame. Then, we apply the estimated offsets and modulation weights to the downsampled input images to generate the interpolated frame at each scale. Finally, the L1 reconstruction losses between the reconstructed frame and the ground truth frame for all scales are combined. Through our experiment, we find that the multi-scale loss does not improve the final results compared to simple L1 reconstruction loss at the finest scale. But we do observe that the multi-scale loss could speed up the convergence during training. For simplicity, all results reported in this paper are obtained by using the simple L1 reconstruction loss at the finest scale.

\paragraph{Training dataset} We use Vimeo-90k training set \cite{xue2019video}, which has 51312 triplets, to train our model. Each triplet has 3 consecutive frames and each frame has a resolution of $448 \times 256$. Horizontal flipping and temporal reversing are adopted as data augmentation.

\paragraph{Training strategy} We train PDWN sequentially. In other words, we first train PDWN without context enhance network for 80 epochs, then finetune the whole system end-to-end for another 20 epochs. We use Adam \cite{kingma2014adam} with $\beta_1 = 0.9$ and $\beta2 = 0.999$ to optimize our model. The initial learning rate is set to 0.0002. Mini-batch size is set to 20. Following the techniques introduced in \cite{niklaus2021revisiting}, we also train a variant of PDWN, called PDWN++, with input normalization, network improvements, and self-ensembling. Specifically, each color channel of the input frames is normalized independently to have zero mean and unit variance. Then, we replace the two-layer convolution with residual blocks. Moreover, the global filter of the deformable convolution that warps frames at level 1 is shared not only between input frames but also across RGB color channels. Finally, 7 transforms, including reverse, flipping, mirroring, reverse and flipping, and rotation by 90, 180, and 270 respectively, are applied during the inference phase for self-ensembling.

\section{Results}
In this section, we first introduce evaluation datasets. Then, we conduct ablation studies to evaluate the contribution of each component and to compare our proposed model with state-of-the-arts on two input frames. Finally, we compare the performance of our models using two vs. four input frames and also compare with other models using four input frames.

\subsection{Evaluation Datasets and Metrics}

\subsubsection{Evaluation Datasets} Our model is trained on a single dataset (Vimeo-90K training set) but validated on multiple datasets including Vimeo-90K\cite{xue2019video} test dataset (448 $\times$ 256), UCF \cite{liu2017video,soomro2012ucf101} dataset (25 FPS, 256 $\times$ 256), and the Middlebury dataset \cite{baker2011database} (typically 640 $\times$ 480). The Middlebury dataset has two subsets. The OTHER set provides the ground-truth middle frames while the EVALUATION set hides the ground-truth and can only be evaluated by uploading the results to the benchmark website.

\subsubsection{Evaluation Metrics} We report PSNR, SSIM \cite{wang2004image}, and Interpolation Error (IE) for model comparison on multiple datasets with various resolutions and contents. IE is the average absolute color error. Higher PSNR or SSIM and lower IE indicate better performance.

\subsection{Ablation Studies}

\subsubsection{Optical flow V.S. DConv} 
\begin{figure*}[tb]
\begin{center}
\includegraphics[width=\linewidth]{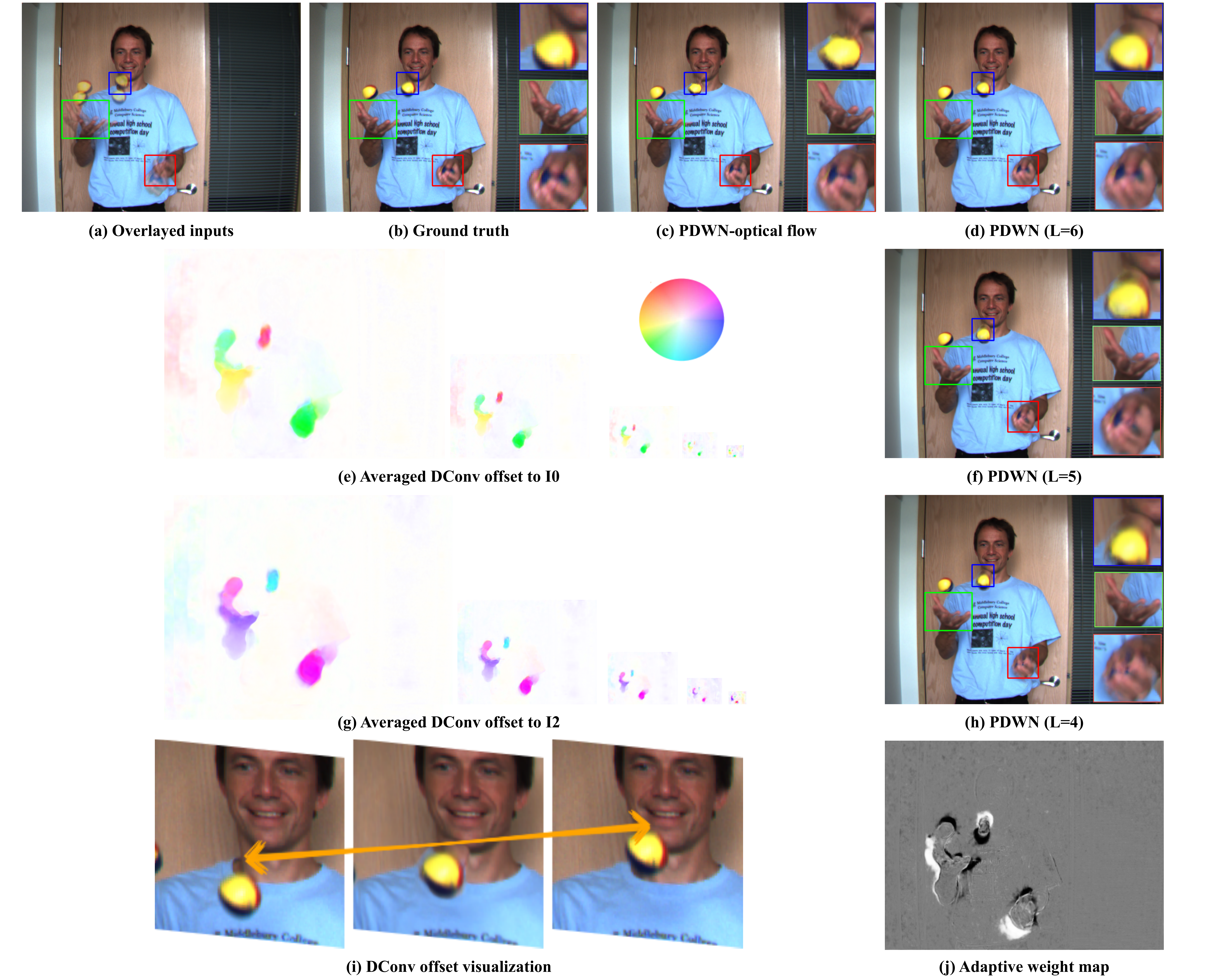}
\end{center}
   \caption{{\bf Analysis on warping operations and network scales \& visualization of DConv offsets and adaptive blending weights.} (c)-(d) Optical flow V.S. DConv. (d), (f), and (h) compares models with different number of scales. The model with larger scales is able to generate more accurate and sharper contents. (i) visualizes the sampling points of DConv in the past and future frames respectively. 
   %Most offsets point to the nearby corresponding positions while some offsets point to places far away from the finger. The synthesis, however, is still better than that using single optical flow, demonstrating the robustness of DConv.  
    (e) and (g) show weighted averaged offsets to the past and future frame respectively at each scale, calculated based on Eq. (\ref{flow}). (j) is the adaptive weight map $\alpha$ for warped past frame, i.e., the weight for the future frame is $1-\alpha$. Thus, the black regions around the hand and ball show PDWN's capacity to handle occlusion.}
\label{fig:flow}
\end{figure*}

To analyze how well the proposed framework performs with different image warping techniques, we train two variants of our approach, one using optical flow and the other using DConv at each scale. To integrate optical flow into our model, PDWN-optical flow generates and refines two sets of optical flow in every pyramid level instead of deformable offsets and modulation weights. Features and frames are backward warped by optical flow in PDWN-optical flow to replace deformable convolution in PDWN. As shown in Table \ref{table:cv} (section 1), DConv outperforms optical flow in terms of all performance metrics, which demonstrates the effectiveness of DConv. In Figure \ref{fig:flow}.(i), we visualize the DConv sampling points in the past and future frame respectively of an occluded point. We observe that the proposed model is able to point to locations in the left frame where the color is similar to the occluded region. As discussed above, DConv offsets can be considered as many-to-one backward warping flow. The redundancy of many-to-one flow makes the model more robust. In Figure \ref{fig:flow}.(e) and \ref{fig:flow}.(g), we visualize the weighted averaged DConv offsets by:
\begin{equation}
    \bar{f}_{1\to i}(x) = \frac{\sum_{j=1}^{|R|} (R(j) + f_{1\to i}(j, x)) m_{1\to i}(j, x)}{\sum_{j=1}^{|R|}m_{1\to i}(j, x)}
    \label{flow}
\end{equation}. 

\begin{table}
\caption{Ablation studies on different components of PDWN}
\begin{center}
\begin{threeparttable}
\begin{tabular}{cccccc}
\toprule
\multirow{2}{*}{Model} & \multicolumn{2}{c}{Vimeo-90k} & \multicolumn{3}{c}{Middlebury OTHER}\\
\cmidrule(r){2-3} \cmidrule(r){4-6}
 & PSNR & SSIM & PSNR & SSIM & IE\\
\midrule
PDWN-optical flow & 34.59 & 0.961 & 35.35 & 0.957 & 2.47\\
\specialrule{0em}{1pt}{1pt}
PDWN w/o modulation  & {35.23} & {0.965} & {\bf37.00} & {\bf0.966} & {2.02} \\
\specialrule{0em}{1pt}{1pt}
PDWN w/ modulation & {\bf 35.38} & {\bf 0.966} & {\bf 37.00} & {\bf 0.966} & {\bf 2.00} \\
\specialrule{0.02em}{1pt}{1pt}
PDWN w/o CV & 35.13 & 0.964 & 37.09 & 0.966 & 1.99 \\
\specialrule{0em}{1pt}{1pt}
PDWN w/ CV & 35.38 & 0.966 & 37.00 & 0.966 & 2.00 \\
\specialrule{0em}{1pt}{1pt}
PDWN w/ learnt CV & {\bf 35.42} & {\bf 0.966} & {\bf 37.17} & {\bf 0.967} & {\bf 1.98}\\
\specialrule{0.02em}{1pt}{1pt}
PDWN w/o coarse-to-fine & 34.54 & 0.959  & 35.95 & 0.961 & 2.19 \\
\specialrule{0em}{1pt}{1pt}
PDWN w/ coarse-to-fine & {\bf 35.42} & {\bf 0.966} & {\bf 37.17} & {\bf 0.967} & {\bf 1.98}\\
\specialrule{0.02em}{1pt}{1pt}
PDWN w/o c. e. & 35.42 & {\bf0.966} & 37.17 & {\bf0.967} & {\bf1.98}\\
\specialrule{0em}{1pt}{1pt}
PDWN w/ c. e. & {\bf 35.44} & {\bf0.966} & {\bf 37.20} & {\bf0.967} & {\bf1.98}\\
\specialrule{0em}{1pt}{1pt}
\bottomrule
\end{tabular}
\begin{tablenotes}
        \footnotesize
        \item[*] CV denotes cost volume and c.e. denotes context enhancement. All models presented here use 6 scales. Models in section 1, 2, and 3 are trained without context enhancement.
      \end{tablenotes}
\end{threeparttable}
\end{center} 
\label{table:cv}
\end{table}
\begin{figure}[ht]
\includegraphics[width=\linewidth]{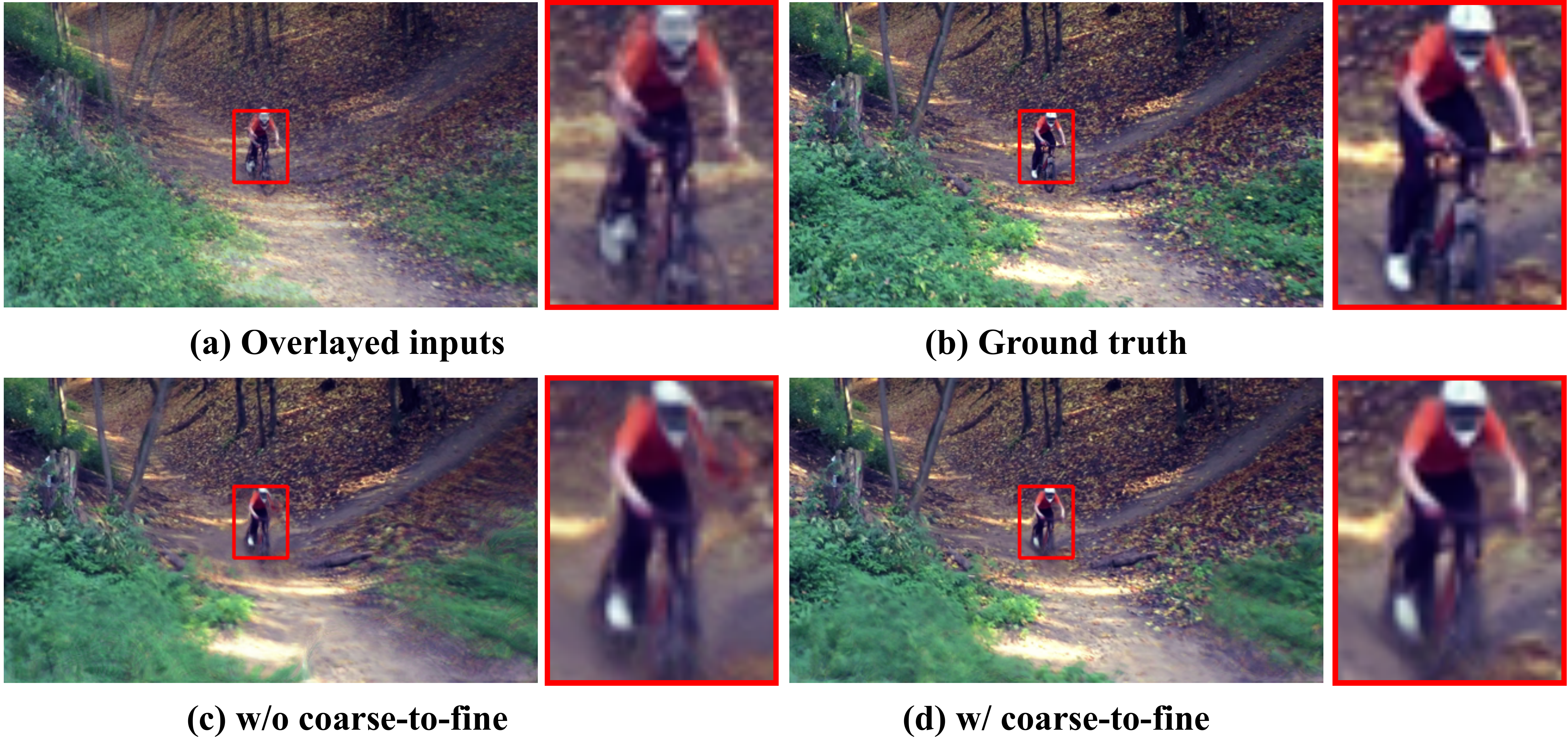}
\caption{{\bf Effect of the coarse-to-fine structure.} By introducing the coarse-to-fine structure, PDWN generates more realistic interpolation results.}
\label{fig:coarse}
\end{figure}
\begin{table}
\caption{Effect of number of scales.}
\label{table:scale}
\begin{threeparttable}
\begin{tabular}{cccccccc}
\toprule
\multirow{2}{*}{Scale}& \multicolumn{1}{c}{Runtime}& \multicolumn{1}{c}{Param.}& \multicolumn{2}{c}{Vimeo-90k} & \multicolumn{3}{c}{Middlebury OTHER} \\
      \cmidrule(r){4-5} \cmidrule(r){6-8}
 & (second) & (million) & PSNR & SSIM & PSNR & SSIM & IE\\
\midrule
L=4 & {\bf 0.0056} & {\bf 1.7} & 35.02 & 0.963 & 36.63 & 0.964 & 2.07\\
\specialrule{0em}{1pt}{1pt}
L=5 & 0.0068 & 3.4 & 35.19 & {\bf0.965} & 36.85 & 0.965 & 2.04\\
\specialrule{0em}{1pt}{1pt}
L=6 & 0.0086 & 6.6 & {\bf 35.23} & {\bf 0.965} & {\bf 37.00} & {\bf 0.966} & {\bf 2.02}\\
\bottomrule
\end{tabular}
\begin{tablenotes}
        \footnotesize
        \item[*] L denotes the number of scales. Note that in this experiment we use a simpler version of DConv where the modulation weights are all set to 1 and the cost volume is predefined. The models trained here are all without context enhancement. The feature size is downsampled 8, 16, 32 times for L = 4, 5, 6, respectively. The runtime is evaluated for interpolating one middle frame of "DogDance" from Middlebury OTHER dataset, with a size of $640\times 480$, on GeForce RTX 2080 Ti.
      \end{tablenotes}
\end{threeparttable}
\end{table}

\begin{figure*}[tb]
\includegraphics[width=\linewidth]{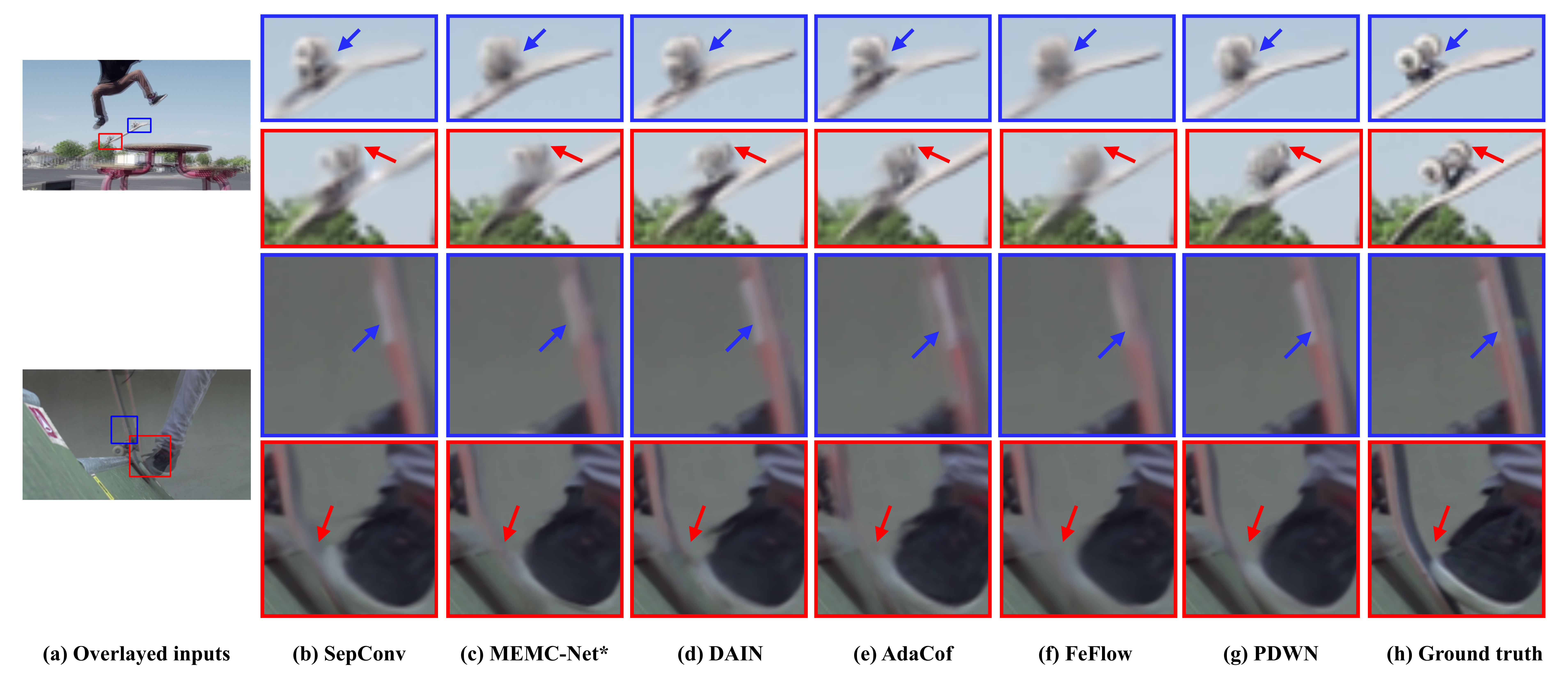}
\caption{{\bf Visualized examples on Vimeo-90k test dataset.}}
\label{fig:vimeo}
\end{figure*}

\begin{table*}[htb]
\caption{Comparison with state-of-the-arts}
\begin{center}
\begin{threeparttable}
\begin{tabular}{lccccccccc}
\toprule
\multirow{2}{*}{Method}&
Runtime& Param.& \multicolumn{2}{c}{Vimeo-90k} & \multicolumn{3}{c}{Middlebury OTHER} & \multicolumn{2}{c}{UCF}\\
        \cmidrule(r){4-5} \cmidrule(r){6-8} \cmidrule(r){9-10}
 & (second) & (million) & PSNR & SSIM & PSNR & SSIM & IE & PSNR & SSIM \\
\midrule
DVF \cite{liu2017video} &-& {\bf\color{red}3.8} & - & - & - & - & - & 34.12 & 0.942\\
\specialrule{0em}{1pt}{1pt}
%SepConv-Lf \cite{niklaus2017video} & {\bf\color{red}0.0032} & 21.6 & 33.47 & 0.952 & 35.12 & 0.954 & 2.44 & 34.56 & 0.945\\
% \specialrule{0em}{1pt}{1pt}
SepConv-L1 \cite{niklaus2017video} & {\bf\color{red}0.0032} & 21.6 & 33.80 & 0.956 & 35.89 & 0.959 & 2.24 & 34.69 & 0.945\\
\specialrule{0em}{1pt}{1pt}
SepConv++ \cite{niklaus2021revisiting} & - & 13.6 & 34.98 & - & \bf\color{blue}{37.47} & - & - & \bf\color{red}{35.29} & - \\ 
\specialrule{0em}{1pt}{1pt}
SuperSlowMo \cite{jiang2018super} & - & 39.6 & - & - & - & - & - & 34.75 & 0.947 \\
\specialrule{0em}{1pt}{1pt}
MEMC-Net* \cite{bao2019memc} & 0.122 & 70.3  & 34.40 & 0.962 & 36.48 & 0.964 & 2.12 & 35.01 & {\bf\color{blue}0.949}\\
\specialrule{0em}{1pt}{1pt}
DAIN \cite{bao2019depth} & 0.125 & 24.0 & 34.71 & 0.964 & 36.70 & 0.965 & 2.04 & 34.99 & {\bf\color{blue}0.949}\\
\specialrule{0em}{1pt}{1pt}
RRPN \cite{zhang2020flexible} & - & - & - & - & - & - & - & 34.76 & - \\ 
\specialrule{0em}{1pt}{1pt}
AdaCof \cite{lee2020adacof} & {\bf\color{blue}0.0043} & 21.8 & 34.35 & 0.956 & 
35.69 & 0.958 & 2.26 & 34.90 & {\bf\color{blue}0.949} \\
\specialrule{0em}{1pt}{1pt}
FeFlow \cite{gui2020featureflow}& 0.7188 & 133.6 & 35.16 & 0.963 & 36.61 & 0.965 & 2.14 & 34.89 & {\bf\color{blue}0.949}\\
\specialrule{0em}{1pt}{1pt}
PDWN (L=6) & 0.0089 & {\bf\color{blue}7.8} & {\bf\color{blue}35.44} & {\bf\color{blue}0.966} & 37.20 & {\bf\color{blue}0.967} & {\bf\color{blue} 1.98} & 35.00 & {\bf\color{red}0.950} \\%34.92 & {\bf \color{blue}0.948}\\
PDWN++ (L=6) & 0.0669 & 8.6 & {\bf\color{red}35.69} & {\bf\color{red}0.968} & {\bf\color{red}38.35} & {\bf\color{red}0.971} & {\bf\color{red} 1.81} & {\bf\color{blue}35.10} & {\bf\color{red}0.950}\\%34.92 & {\bf \color{blue}0.948}\\
\bottomrule

\end{tabular}
    \begin{tablenotes}
        \footnotesize
        \item[*]  PDWN achieves on par performance with much fewer parameters compared to previous methods.
        \item[*]  The runtime of DAIN and MEMC-Net* is reported in their paper on a 640x480 image using an NVIDIA Titan X (Pascal) GPU card. Other runtime numbers reported are estimated for "DogDance" image on an Nvidia RTX 2080 Ti GPU card.
        \item[*] The number in {\color{red}red} and {\color{blue}blue} represents the best and second best performance.
     \end{tablenotes}
 \end{threeparttable}
\end{center}
\label{table:vimeo}
\end{table*}

\begin{figure*}[htb]
\includegraphics[width=\linewidth]{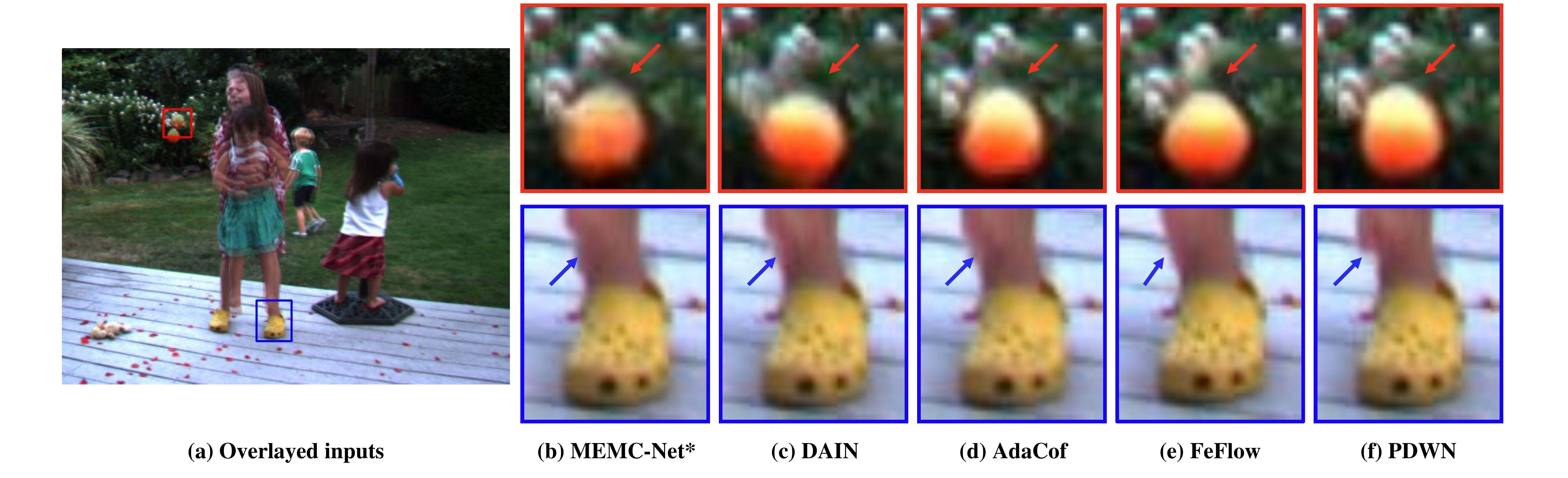}
\caption{{\bf Visualized examples on Middlebury EVALUATION dataset.} PDWN generates high-quality details on the foot and girl's toe while other methods produce blurry output. Moreover, PDWN shows its capacity in dealing with occlusion and semantic shape distortion on the ball and white flowers.}
\label{fig:middle}
\end{figure*}

\begin{table*}[htb]
\caption{Results on Middlebury EVALUATION datase}

\begin{center}
\renewcommand\tabcolsep{2.5pt}
% \resizebox{\textwidth}{!}{
\begin{threeparttable}
\begin{tabular}{lcc|cccccccccccccccc}
\toprule
\multirow{2}{*}{Method} & \multicolumn{2}{c}{Average} & \multicolumn{2}{c}{Mequon} & \multicolumn{2}{c}{Schefflera} & \multicolumn{2}{c}{Urban} &  \multicolumn{2}{c}{Teddy} & \multicolumn{2}{c}{Backyard} & \multicolumn{2}{c}{Basketball} & \multicolumn{2}{c}{Dumptruck} & \multicolumn{2}{c}{Evergreen}\\
     \cmidrule(r){2-3} \cmidrule(r){4-5} \cmidrule(r){6-7} \cmidrule(r){8-9} \cmidrule(r){10-11} \cmidrule(r){12-13} \cmidrule(r){14-15}\cmidrule(r){16-17}\cmidrule(r){18-19}
 & IE & NIE & IE & NIE & IE & NIE & IE & NIE & IE & NIE & IE & NIE & IE & NIE & IE & NIE & IE & NIE\\
\midrule
%SepConv-L1\cite{niklaus2017video} & 5.60 &  &2.52 & & 3.56 & & 4.17 & & 5.41 & & 10.2 & & 5.47 & & 6.88 &  & 6.63 & \\
%\specialrule{0em}{1pt}{1pt}
SuperSlowMo\cite{jiang2018super} & 5.31 &  0.78 &2.51 & 0.59 & 3.66 &  0.72 & 2.91 & 0.74 & 5.05 & 0.98 & 9.56 & 0.94 & 5.37 & 0.96 & 6.69 &  0.60 & 6.73 & 0.69\\
\specialrule{0em}{1pt}{1pt}
MEMC-Net*\cite{bao2019memc} & 4.99 & 0.74 & 2.39 & 0.59 & 3.36 & 0.64 & 3.37 &  0.80 & 4.84 &  0.88 & 8.55 & 0.88 & 4.70 & {\bf \color{blue}0.85} & 6.40 &  0.64 & 6.37 & 0.63\\
\specialrule{0em}{1pt}{1pt}
DAIN\cite{bao2019depth}& 4.85 &  {\bf \color{blue}0.71} & 2.38 &  0.58 & 3.28 &  0.60 & 3.32 &  {\bf\color{blue}0.69} & 4.65 &  {\bf\color{blue}0.86} & 7.88 &  0.87 & 4.73 &  {\bf \color{blue}0.85} & 6.36 & {\bf\color{blue}0.59} & 6.25 & 0.66\\
\specialrule{0em}{1pt}{1pt}
AdaCof\cite{lee2020adacof} & 4.75 & 0.73 & 2.41 & 0.60 & {\bf\color{blue}3.10} & 0.59 & 3.48 & 0.84 & 4.84 & 0.92 & 8.68 & 0.90 & {\bf \color{blue}4.13} & {\bf\color{red} 0.84} & {\bf \color{blue}5.77} & {\bf\color{red}0.58} &	{\bf \color{red}5.60} & {\bf\color{red}0.57} \\
\specialrule{0em}{1pt}{1pt}
FeFlow\cite{gui2020featureflow} & 4.82 &  {\bf \color{blue}0.71} &  {\bf \color{blue}2.28} & {\bf\color{blue}0.51} & 3.50 &  0.66 & {\bf \color{blue}2.82}  & 0.70 & 4.75 &  0.87 & {\bf \color{red}7.62} &  {\bf\color{red}0.84} & 4.74 &  0.86 & 6.07 &  0.64 & 6.78 & 0.67 \\
\specialrule{0em}{1pt}{1pt}
RRPN \cite{zhang2020flexible} & 4.93 & 0.75 & 2.38 & 0.53 & 3.70 & 0.69 & 3.29 & 0.87 & 5.05 & 0.94 & 8.20 & 0.88 & 4.38 & 0.88 & 6.50 & 0.65 & 6.00 & 0.62\\
\specialrule{0em}{1pt}{1pt}
SepConv++ \cite{niklaus2021revisiting} & {\bf \color{red}3.88} & 0.73 & 2.39 & 0.58 & {\bf\color{red}2.98} & {\bf\color{red}0.56} & 3.34 & 0.95 & {\bf\color{blue}4.49} &0.87& {\bf\color{blue}7.64} & {\bf\color{blue}0.85}& {\bf\color{red}3.77} & {\bf\color{red}0.84} & {\bf\color{red}5.26} & {\bf\color{blue}0.59} & {\bf\color{blue}5.71} & {\bf\color{blue}0.59}\\
\specialrule{0em}{1pt}{1pt}
PDWN (L=6) & {\bf \color{blue}4.71} & {\bf \color{red} 0.69} &{\bf \color{red}2.09} & {\bf\color{red}0.46} & 3.12 & {\bf\color{blue} 0.58} & {\bf \color{red} 2.38} & {\bf\color{red}0.64} & {\bf \color{red} 4.29} & {\bf\color{red}0.85} & 8.61 & 0.87 & 4.80 & 0.88 & 6.24 & 0.60 &	6.18 & 0.62\\
\bottomrule
\end{tabular}

\begin{tablenotes}
\item[*] NIE: normalized interpolation error.
\end{tablenotes}

\end{threeparttable}
% }
\end{center}
\label{table:middle}
\end{table*}

\begin{figure*}[htb]
\includegraphics[width=\linewidth]{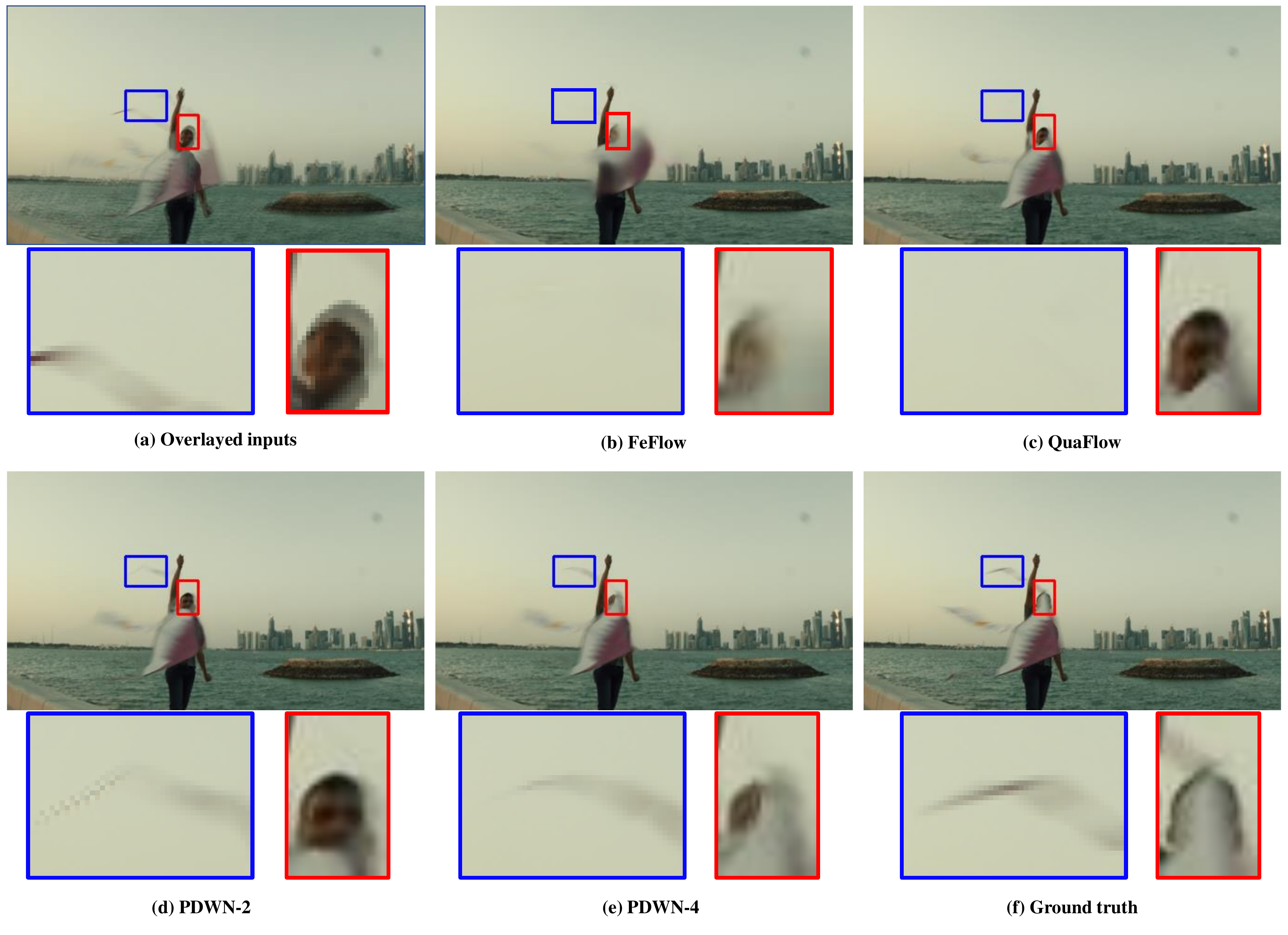}
\caption{{\bf Visualized examples of the extended PDWN with 4 input frames.} PDWN-2 takes only 1 past frame and 1 future frame as input. QuaFlow and PDWN-4 take 2 past frames and 2 future frames as input.}
\label{fig:vimeo177}
\end{figure*}

\begin{table}[htb]
\caption{Results of the extended PDWN with four input frames on Vimeo-90K spetulet dataset}
\begin{center}
\begin{threeparttable}
\begin{tabular}{lcccc}
\toprule
\multirow{2}{*}{Method}&
Runtime& Param.& \multicolumn{2}{c}{Vimeo-septuplet}\\
        \cmidrule(r){4-5} 
 & (second) & (million) & PSNR & SSIM\\
\midrule
FeFlow \cite{gui2020featureflow} & 0.221 & 133.6 & 33.88 & 0.946 \\
\specialrule{0em}{1pt}{1pt}
PDWN-2 & {\bf 0.010} & {\bf 7.4} & 35.53 & 0.958\\
\specialrule{0em}{1pt}{1pt}
QuaFlow \cite{xu2019quadratic}  & 0.090 & 19.6 & 34.28 & 0.950\\
\specialrule{0em}{1pt}{1pt}
PDWN-4 & 0.012 & 8.3  & {\bf 35.93} & {\bf 0.960}\\

\bottomrule
\end{tabular}
    \begin{tablenotes}
        \footnotesize
        \item[*]  FeFlow and PDWN-2 take only 1 past frame and 1 future frame as input. QuaFlow and PDWN-4 take 2 past frames and 2 future frames as input.
        \item[*] Both PDWN-2 and PDWN-4 have 6 pyramid levels and no contexual enhancement module.
        \item[*] The runtime reported is the average runtime for Vimeo-septuplet dataset with image size $448\times256$ on an Nvidia Tesla V100 GPU card.
     \end{tablenotes}
 \end{threeparttable}
\end{center}
\label{table:4input}
\end{table}

\subsubsection{Cost volume} 
To analyze the effectiveness of using cost volumes, we consider three variants of our approach. The first model takes warped features only as input to the first conv layer in the offset estimator in Figure \ref{fig:network_structure}.(b). The second model first computes the cost volume between two warped features, then concatenates the cost volume and the original features to estimate DConv offset residuals. The third model replaces the cost volume layer with a two-layer CNN to learn the matching cost between two warped features. As shown in Table \ref{table:cv} (section 2), cost volumes bring additional improvements without adding more parameters on Vimeo-90K dataset. Replacing the predefined cost with the learnt cost further improves the results for both datasets.

\subsubsection{Coarse-to-fine successive refinement manner} 
In the proposed model, we warp features and construct the matching cost between warped features to estimate DConv offset {\it residuals} $\Delta f_{1\to i}^l$ at every pyramid level in a coarse-to-fine manner. It reduces the distance between two input frames gradually and is particularly important when the ground truth motion is large. We investigate the contribution of this coarse-to-fine structure via training another variant of our model, without the coarse-to-fine structure. In other words, this model is simply a UNet structure with 6 spatial scales that takes two images $I_0$ and $I_2$ as input and directly outputs DConv offsets and modulation weights in the finest scale. We show the quantitative results in Table \ref{table:cv} (section 3) and qualitative results in Figure \ref{fig:coarse}. By introducing the coarse-to-fine structure, the performance is significantly improved, demonstrating the effectiveness of our successive coarse-to-fine successive refinement approach.

\subsubsection{Impact of the number of scales}
To analyze the impact of the number of scales on the performance, we investigate three different pyramid scales ($L$ = 4, 5, 6). Quantitative results are shown in Table \ref{table:scale}, and the visual comparison is provided in Figure \ref{fig:flow}. We find that with model size increasing from 1.7, 3.4, to 6.6 million, the PSNR steadily get better from 36.63, 36.85, to 37.00 dB on Middlebury OTHER dataset. The example in Figure \ref{fig:flow} also shows that the model using more scales generates sharper outcomes. The gain on Vimeo-90K, however, is not as significant as that on Middlebury OTHER dataset. That is probably because Middlebury OTHER dataset has a larger image size (and hence larger motion in terms of pixels) than Vimeo-90K dataset. Even though the model size almost doubles with each additional scale, the runtime only increases slightly, as the lower scale images and features have a smaller spatial dimension. 

\subsubsection{Adaptive Blending Weight} Figure \ref{fig:flow}.(j) shows an example of adaptive blending weight map. As discussed in section 3.3, $\alpha(x) = 0$ means pixel  x from $I_0$ is occluded and pixels x from $I_1$ is fully trusted. The black region around the ball in the weight map indicates that our model can detect and solve occlusion by selecting pixels from the previous and following frames softly.

\subsubsection{Context enhancement network}
To analyze the contribution of the context enhancement module, we train a variant of PDWN without context enhancement and show the results in Table \ref{table:cv}. Though DAIN gains significantly from adding the context enhancement module (0.27 dB on Vimeo-90k in terms of PSNR) \cite{bao2019depth}, 
the context enhancement network has little contribution to PDWN. By adding the context enhancement network, the number of model parameters increases from 7.4 million to 7.8 million and the runtime increases from 0.0082 to 0.0086 for interpolating "DogDance" image (640$\times$480) in Middlebury-OTHER dataset, using an NVIDIA RTX 8000 GPU card.

\subsection{Comparison with state-of-the-arts}
We compare our model with state-of-the-art video interpolation models both quantitatively and qualitatively, including deep voxel flow (DVF) \cite{liu2017video},  SepConv\cite{niklaus2017video}, SepConv++ \cite{niklaus2021revisiting}, SuperSloMo \cite{jiang2018super}, MEMC-Net* \cite{bao2019memc}, DAIN \cite{bao2019depth}, AdaCof \cite{lee2020adacof}, FeFlow \cite{gui2020featureflow}, on three different datasets, Vimeo-90K, UCF, and Middlebury dataset. Note that we only compare with methods which use backward optical flow or DConv for backward image warping. For SepConv, AdaCof, and FeFlow, we download their published models and test on the testing datasets. For DVF, SuperSloMo, MEMC-Net*, and DAIN, we calculate the numbers from their published interpolated data. For RRPN and SepConv++, we directly report their published numbers.

As shown in Table \ref{table:vimeo}, our proposed model outperforms all methods on Vimeo-90k dataset and Middlebury OTHER dataset except SepConv++. Using similar techniques applied to SepConv++, PDWN++ surpasses SepConv++ for 0.88 dB on Middlebury OTHER dataset with respect to PSNR. Meanwhile, the number of model parameters increases from 7.8 million to 8.6 million and the runtime increases nearly 8 times. On UCF dataset, our model achieves on par performance with state-of-the-art methods. Note that DAIN uses additional depth information to detect occlusion in order to compensate errors in the linear interpolated optical flow. DAIN relies on the accuracy of depth information, i.e., their model cannot learn meaningful depth information without a good initialization of (pretrained) depth estimation network and thus yields lower quality results than MEMC-Net. Our model does not need depth information for training information but still achieve 0.73 dB higher PSNR than DAIN on Vimeo-90K. FeFlow uses multiple groups of DConv offsets in every layer to avoide occlusion and edge maps generated by BDCN \cite{He_2019} as structure guidance. Compared to FeFlow, our model performs better on Vimeo-90K without edge maps and with only a single group of DConv offsets, which demonstrated the supremacy of using DConv in a coarse-to-fine manner. Moreover, our model size is only 5.8\% of that of FeFlow. Figure \ref{fig:vimeo} presents two examples from Vimeo-90k dataset. Notably, our model generates the sharpest results among all compared methods. %SoftSplate relies on ground truth optical flow to train the optical flow estimator while our model does not rely on any additional information. Therefore, in cases where the ground truth optical flow is hard to obtained and hard to generalize from other existing flow datasets, our model might be preferred over SoftSplate. 

Table \ref{table:middle} shows the comparison on Middlebury EVALUATION dataset. Our proposed method performs favorably against state-of-the-art methods. Our model performs well quantitatively on sequences with small motion or fine textures such as {\it Mequon, Teddy} and {\it Schefflera}. For videos with complicated motions, Figure \ref{fig:middle} shows a visualized example. Our model produces more details at the girl's toe in the {\it backyard} example while other methods output blurry results. And our model handles occlusion well around the boundary of the orange ball.

\subsection{Extending to four input frames} 
Vimeo-90K septuplet dataset is used to train and test our extended model PDWN-4 which takes four input frames as input and has 6 pyramid levels. We use frame 1, 3, 5, and 7 to interpolate frame 4 and compare the interpolated frame 4 with the original frame 4 for every sequence in Vimeo-90K septuplet dataset. We compare the results with our two-input model PDWN-2 and state-of-the-art methods including FeFlow \cite{gui2020featureflow} and QuaFlow \cite{xu2019quadratic}. PDWN-2 is pretrained on Vimeo-90K triplet dataset and finetuned on Vimeo-90K septuplet dataset. Results are given in Table \ref{table:4input}. Figure \ref{fig:vimeo177} shows visualized results on the Vimeo-90K septuplet test dataset. Both the quantitative and visual evaluations demonstrate that the extended PDWN with four input frames can significantly improve the interpolation accuracy over using two input frames, with only modest increases in the model size and the runtime. Furthermore, both PDWN-2 and PDWN-4 yield better results than QuaFlow that uses four input frames.

\section{Conclusion}

%In this work, we propose a video interpolation model that estimates the many-to-one flows of the middle frame to the left and right input frames. We show that estimating the many-to-one flows together with modulation maps is more robust than estimating the optical flow. The offset estimator can benefit from using the cost volumes computed from the aligned features, compared to using the aligned features directly. Our model is significantly smaller in model size and requires substantially less inference time compared to state-of-the-art models and yet achieves better or on-par interpolation accuracy. Besides, our model does not rely on additional information (e.g. ground truth depth information or optical flow) for training. Moreover, our model that uses two input frames can be extended to using four input frames relatively easily, with only a small increase in the model size and the inference time, and yet the extended model significantly improves the interpolation accuracy.
In this work, we propose a pyramid video interpolation model that estimates the many-to-one flows with modulation maps of the middle frame to the left and right input frames. We show that the offset estimator can benefit from using the cost volumes computed from the aligned features, compared to using the aligned features directly. Our model is significantly smaller in model size and requires substantially less inference time compared to state-of-the-art models and yet achieves better or on-par interpolation accuracy. Besides, our model does not rely on additional information (e.g. ground truth depth information or optical flow) for training. Moreover, our model that uses two input frames can be extended to use four input frames easily, with only a small increase in the model size and the inference time, and yet the extended model significantly improves the interpolation accuracy.

A recent work \cite{Niklaus_CVPR_2020}, which proposes a differentiable forward warping operation using forward optical flow to handle occlusion and dis-occlusion regions directly, outperforms all backward-flow-based methods. It shows a promising direction for video interpolation. In future work, we will also explore how to combine forward warping with a coarse-to-fine structure. Furthermore, we will explore the integration of PDWN in video coding, where the encoder can encode every other frame; Skipped frames will be interpolated by the PDWN method and the interpolation error images can be additionally coded. 

\clearpage
\clearpage
\printbibliography

\end{document}